\documentclass[letterpaper, 10 pt, conference]{ieeeconf}  %

\IEEEoverridecommandlockouts                              %

\usepackage{cite}
\usepackage{graphicx}
\usepackage{booktabs}
\usepackage{multirow}
\usepackage{makecell}
\usepackage{xcolor}
\usepackage{colortbl}
\usepackage{amsmath}
\usepackage{amssymb}
\usepackage{hyperref}
\usepackage[font=small,labelfont=bf,tableposition=top]{caption}
\usepackage{verbatim}

\hypersetup{
    pdfauthor={Jen-Wei Wang, Sarthak Kaingade, Andrea Tagliabue, Nicholas Morozovsky},
    pdftitle={X-OP: Cross-Morphology Whole-Body Teleoperation via MPC Retargeting},
    pdfsubject={Robotics, Whole-Body Teleoperation, Model Predictive Control},
    pdfkeywords={whole-body teleoperation, MPC, motion retargeting, humanoid, mobile manipulator, XR, MuJoCo, MPPI}
}

\title{\LARGE \bf
X-OP: Cross-Morphology Whole-Body Teleoperation \\ via MPC Retargeting
}

\author{
    Jen-Wei Wang$^{1,2}$, 
    Sarthak Kaingade$^{1}$,
    Andrea Tagliabue$^{1}$,
    and Nicholas Morozovsky$^{1}$\\
    $^{1}$Amazon, 
    $^{2}$University of California, Berkeley
}

\begin{document}

\twocolumn[{
    \renewcommand\twocolumn[1][]{#1} 
    \maketitle
    
    \begin{center}
        \includegraphics[width=0.9\textwidth]{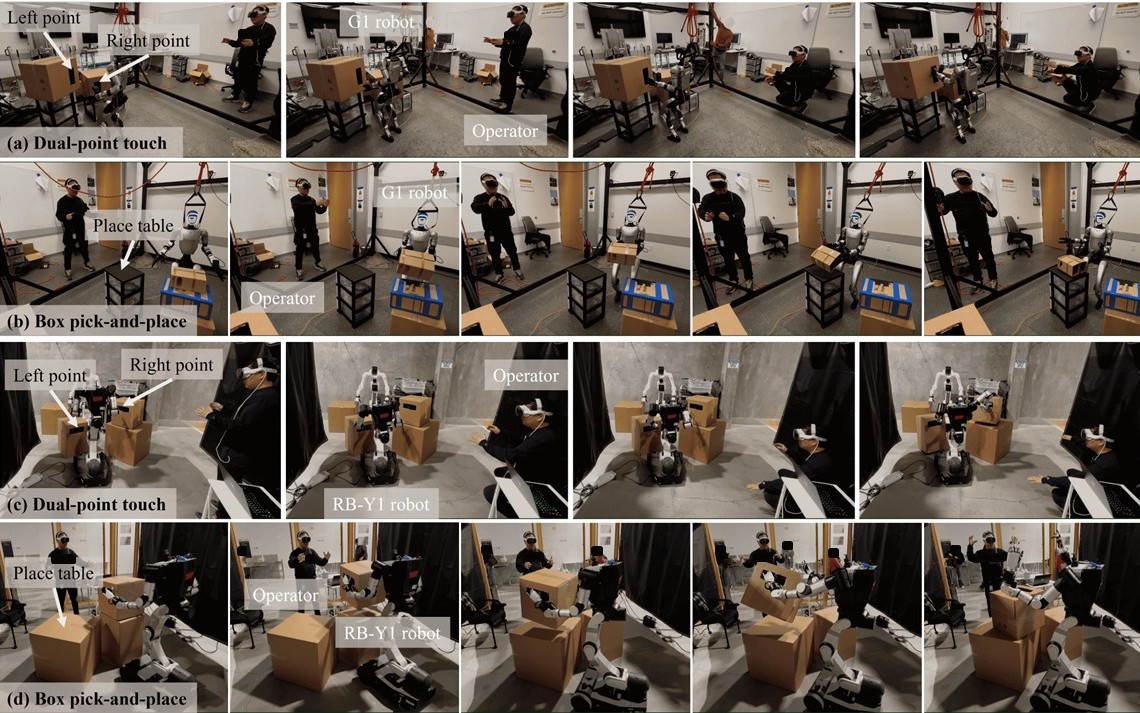}
        \captionof{figure}{Each row shows a teleoperation sequence on either the Unitree G1 humanoid or the Rainbow RB-Y1 mobile manipulator, performing dual-point touch (touching two holes on separate boxes) or box pick-and-place (transferring a box to a target table).}
        \label{fig:firstfigure}
    \end{center}
    
}]

\begin{abstract}

Whole-body teleoperation is essential for scalable robot data collection in loco-manipulation tasks, yet existing approaches relying on exoskeleton suits or multi-camera setups impose prohibitive cost, complexity, and environmental constraints. Recent methods employing single extended reality (XR) devices with end-to-end reinforcement learning policies partially address these limitations but require robot-specific retraining, suffer from out-of-distribution failures, and depend on motion retargeting that neglects dynamic feasibility. To overcome these challenges, we propose a hierarchical whole-body teleoperation framework driven by a single XR device that generalizes across diverse robot morphologies without retraining robot-specific policies. At its core, the framework employs a Model Predictive Control (MPC)-based motion retargeter that jointly optimizes alignment with the operator's intent and the robot's dynamic feasibility, generating optimal commands for existing low-level controllers. To ensure robust online execution, we introduce a state synchronization method that properly resets the simulator state at each MPC step to handle noisy real-world measurements and contact sensitivity, and integrate SLAM-based global pose feedback to mitigate drift during long-term teleoperation. Simulation results demonstrate that our method achieves higher success rates across whole-body control tasks for both a humanoid—with over 30\% reduced completion time and 20\% lower power consumption—and a mobile manipulator—with zero collisions—compared to baselines. Real-world experiments further validate the effectiveness and flexibility of our method, demonstrating the successful deployment of the proposed retargeter on both platforms for whole-body control tasks and the ease of allowing users to adjust teleoperation behavior based on their preferences. This plug-and-play framework offers a scalable, flexible, and morphology-agnostic solution for whole-body robot teleoperation, enabling real-time behavioral customization and broad applicability across platforms.

\end{abstract}

\begin{figure*}[t]
    \vspace{2mm}
    \centering
    \includegraphics[width=0.9\textwidth]{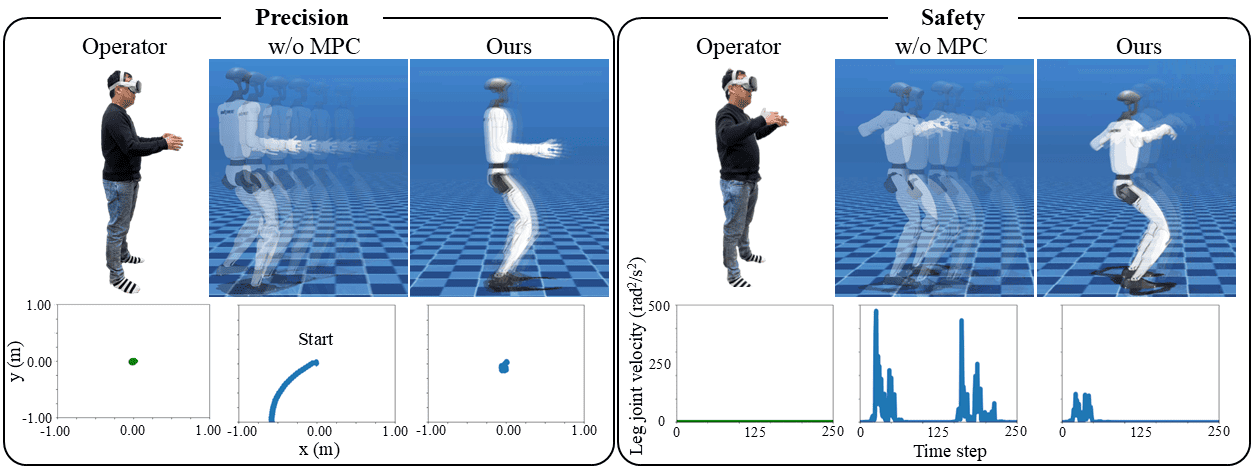}
    \caption{Our framework achieves superior precision and safety over methods without MPC. Such methods directly transform human motion to robot motion using heuristic rules in an open-loop manner, causing gradual drift even when the operator is stationary, whereas our framework maintains reliable positioning. For safety, when the operator commands destabilizing upper-body configurations, our framework prevents falls by proactively squatting and rotating the torso, deviating slightly from the intended pose but keeping the robot stable.}
    \label{fig:precision_safety}
\end{figure*}

\section{INTRODUCTION}

Many manipulation tasks in both domestic and industrial environments require whole-body coordination skills. For instance, relocating an object placed on the ground involves squatting to grasp it, standing while maintaining hold, navigating to a new location, and placing it on a surface. Enabling such loco-manipulation capabilities in robots, whether humanoids or mobile manipulators, has become a key research objective. Consequently, numerous robot learning algorithms aim to achieve this by training foundation models on large-scale real-world data. Efficient data collection for this objective necessitates a whole-body teleoperation system that is easy to use, generalizable across robot morphologies, and adaptable to varying tasks and user preferences. Prior teleoperation methods typically rely on exoskeleton suits \cite{ben2025homie} or multi-camera setups \cite{ze2025twist} such as \href{https://optitrack.com/}{OptiTrack} or \href{https://www.vicon.com/}{VICON} \cite{ze2025twist}, both of which impose complex and restrictive requirements. Specifically, exoskeleton suits are bulky and expensive, making data collection difficult to scale and requiring operators to have large workspace for maneuvering. Multi-camera setups, on the other hand, demand careful camera placement and calibration prior to operation, and constrain the operator to work strictly within the cameras' field of view.

\begin{figure*}[t]
    \vspace{2mm}
    \centering
    \includegraphics[width=0.8\textwidth]{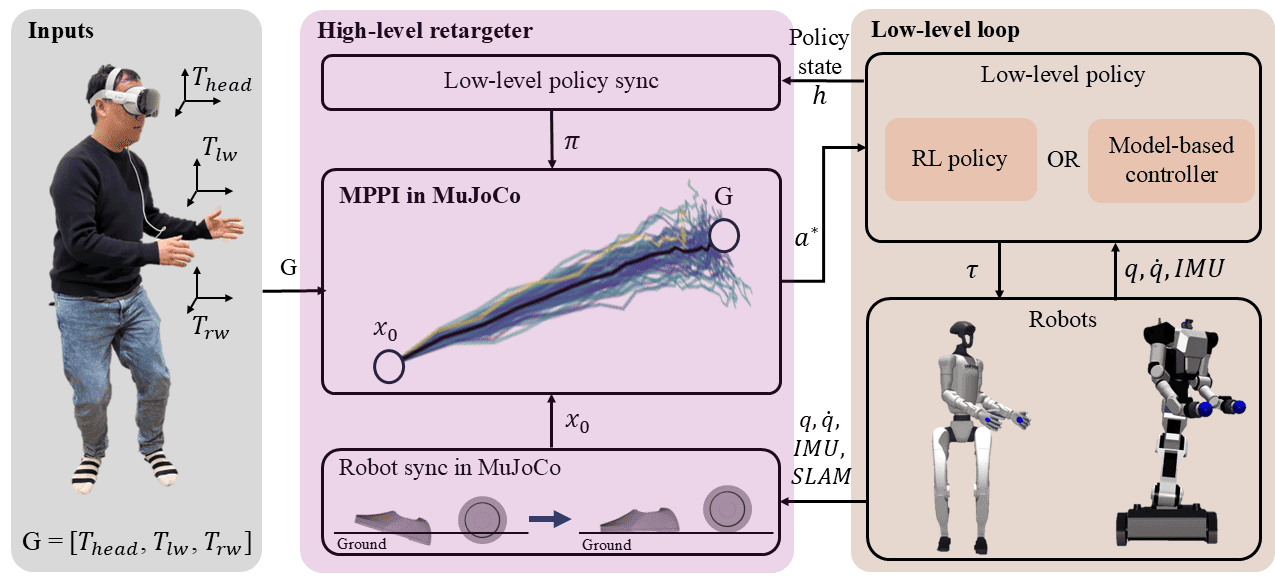}
    \caption{Overview of our hierarchical whole-body teleoperation framework. An operator wearing an XR device streams headset, left-wrist, and right-wrist target poses to the MPC-based retargeter, which uses a closed-loop MuJoCo simulation with the low-level policy as its dynamics model. The retargeter optimizes commands $a^*$ balancing goal alignment with safety. An optimization-based reset procedure synchronizes the simulation state at each MPC step. The framework is platform-agnostic, supporting arbitrary low-level policies and robot morphologies.}
    \label{fig:framework}
\end{figure*}

Recent works \cite{li2025clone, li2025amo, seo2025legato} adopt a single extended reality (XR) device—a headset equipped with built-in cameras and inertial sensors for spatial tracking—that operates in diverse environments with minimal setup. Since a single XR headset cannot directly estimate foot placements, posing a significant challenge for whole-body teleoperation, these methods train end-to-end reinforcement learning (RL) policies for whole-body control, enabling a robot to track end-effector poses streamed from the XR device—such as headset, left wrist, and right wrist poses—while simultaneously coordinating lower-body movements to maintain balance and locomotion. Training such policies typically requires a human motion library \cite{mahmood2019amass}, where motions captured from humans must be retargeted from human morphology to robot morphology using existing retargeters \cite{darvish2019whole, araujo2025retargeting}. However, these retargeters primarily focus on preserving geometric similarity such as joint angles and end-effector positions, without considering whether the resulting motions are dynamically feasible for the robot, and potentially leading to physically unrealizable motions for RL policies. Beyond this challenge, these approaches have several notable limitations: they require retraining robot-specific policies for teleoperation, may fail on out-of-distribution (OOD) targets, leading to imprecise and unstable control (Figure \ref{fig:precision_safety}), and preclude real-time behavioral adjustments due to their offline training paradigm.

To address these challenges, we propose a hierarchical whole-body teleoperation framework (Figure \ref{fig:framework}) that generalizes across diverse robot morphologies by leveraging existing low-level policies (e.g., model-based controllers or RL policies) not originally developed for teleoperation. Rather than learning an end-to-end model to track teleoperation targets, we employ a Model Predictive Control (MPC)-based retargeter that generates optimal commands for the low-level policy by jointly considering alignment with the operator's motions and the robot's dynamic feasibility. Because low-level policies accepting high-level commands—such as desired base linear velocity or head height—are now widely available \cite{li2025amo, zhang2025falcon}, our framework serves as a plug-and-play module compatible with various controllers and robot platforms. Moreover, since MPC explicitly optimizes actions online over cost functions and constraints, both the formulation and parameters can be adjusted in real time, enabling flexible and customizable teleoperation. To solve online the optimization problem associated with MPC, we employ Model Predictive Path Integral (MPPI) \cite{williams2017information} with a digital twin comprising the MuJoCo simulator, a model of the robot, and the low-level policy to predict future rollouts. Given that real-world measurements are noisy and rollout predictions are sensitive to robot-ground contact, we propose a state synchronization method that properly resets the simulator state at each MPC step.

In both simulation and real-world experiments, we show that our approach outperforms baselines in whole-body control tasks, achieving higher success rates while significantly reducing completion time by over 30\% and power consumption by over 20\% on a humanoid, and avoiding all collisions on a mobile manipulator. We demonstrate successful teleoperation on two robotic platforms for whole-body control tasks in real-world, and also highlight its flexibility, allowing users to choose between single- and multi-target-pose alignment, incorporate additional constraints, and tune the aggressiveness or conservativeness of the retargeted motion. The ablation study confirms that the proposed state synchronization method is essential for ensuring reliable ground contact conditions in MPPI when relying on noisy sensor inputs such as SLAM, IMU, and motor encoders. 

In summary, our \textbf{contributions} are:
\begin{itemize}
    \item A hierarchical whole-body teleoperation framework driven by a single XR device that decouples retargeting from low-level control loop, enabling generalization across diverse robot morphologies by aligning the retargeter's output with the command interface of off-the-shelf low-level policies without retraining.
    \item An MPC-based motion retargeter with a state synchronization method that enables precise and safe teleoperation by jointly optimizing operator intent tracking and dynamic feasibility.
    \item Comprehensive simulation and real-world experiments under realistic sensing on a humanoid robot and a mobile manipulator, validating the effectiveness and versatility of our framework across various whole-body control tasks.
\end{itemize}

\section{Related Works}

Existing whole-body teleoperation systems rely on exoskeleton suits \cite{ben2025homie}, motion capture systems \cite{ze2025twist}, or mixed device combinations \cite{lu2025mobile}, all requiring elaborate setup and restrictive environments. Single RGB camera methods \cite{he2025asap, he2024learning} reduce hardware needs but suffer from occlusion artifacts. Single XR device approaches \cite{li2025clone, li2025amo} offer portability, yet require robot-specific retraining and lack guarantees on tracking precision or safety. Universal frameworks \cite{he2024omnih2o} accept arbitrary inputs but still require retraining for each robot morphology. In contrast, our modular framework supports plug-and-play integration across input devices and robot platforms while explicitly enforcing precision and safety without retraining. For motion retargeting, it transforms human motion into corresponding robot motion. End-effector-centric methods \cite{seo2025legato} use whole-body inverse kinematics for joint position tracking, while end-to-end approaches \cite{darvish2019whole, araujo2025retargeting} transfer human motion libraries into robot trajectories as reference signals for reinforcement learning. These methods primarily adopt a geometric perspective—defining morphing functions or learned mappings between kinematic structures—but neglect dynamic feasibility of the retargeted motions. In contrast, our retargeting formulation accounts for the robot's dynamics, solves the mapping online, and jointly enforces safety constraints and motion alignment with the operator.

\section{MPC-based Retargeter}

This section presents the proposed MPC-based formulation for whole-body teleoperation. We model teleoperation as a goal-reaching problem, where the operator continuously specifies time-varying goals during operation. Unlike previous approaches \cite{ze2025twist, li2025amo, li2025clone} that directly map operator motions to the command space via hand-designed retargeting functions, our method solves an MPC optimization problem that explicitly accounts for the closed-loop dynamics of the whole-body policy and robotic system. Through appropriate cost function design, the MPC generates optimal commands that balance operator motion alignment with closed-loop system safety.

Figure \ref{fig:framework} illustrates the proposed hierarchical framework. The MPC-retargeter serves as a high-level controller that maps operator motions to the command space of a low-level policy, which in turn generates motor torques. Notably, the low-level policy can be either a learning-based policy or a traditional model-based controller. Since mature whole-body controllers with basic functionalities (e.g., velocity tracking or body pose tracking) are readily available across various platforms—such as whole-body RL policies for humanoid robots \cite{cheng2024expressive, li2025amo, zhang2025falcon}—our framework can leverage these existing policies in the teleoperation system.

\subsection{Problem Formulation}
\label{subsec:problem_formulation}

\begin{subequations}
\label{eq:hierarchical_mpc}
\begin{align}
\min_{a_0, \ldots, a_{H-1}} \quad & \sum_{k=0}^{H-1} l(x_k, a_k, G(t)) + V_f(x_H, G(t)) \label{eq:hmpc_cost}\\
\textrm{s.t.} \quad & \tau_k = \pi(x_k, a_k, h_k), \quad k = 0, \ldots, H-1 \label{eq:hmpc_policy}\\
& x_{k+1} = g(x_k, \tau_k), \quad k = 0, \ldots, H-1 \label{eq:hmpc_simulator}\\
& a_k \in \mathcal{A}, \quad k = 0, \ldots, H-1 \label{eq:hmpc_action_constraints}\\
& x_0 = x(t) \label{eq:hmpc_initial_robot} \\
& h_0 = h(t) \label{eq:hmpc_initial_policy}
\end{align}
\end{subequations}

where $a_k$ denotes the high-level action generated by the MPC at time step $k$, and $\mathcal{A} = \{a \mid a_{\min} \leq a \leq a_{\max}\}$ represents the admissible action set defined by the lower bound $a_{\min}$ and upper bound $a_{\max}$. $x_k$ denotes current state consisting of robot base position $p \in \mathbb{R}^3$, robot base orientation $\theta \in \mathbb{R}^3$, robot joint position $q \in \mathbb{R}^n$, robot base linear velocity $v \in \mathbb{R}^3$, robot base angular velocity $\omega \in \mathbb{R}^3$ and joint velocity $\dot{q} \in \mathbb{R}^n$. The function $\pi(\cdot, \cdot)$ represents the low-level policy that maps the current state, high-level action and hidden parameter $h_k$ to joint torques $\tau_k$. In RL policy, the hidden parameter $h_k$ can be the state-action history. The function $g(\cdot, \cdot)$ corresponds to the simulator dynamics, which computes the next state $x_{k+1}$ given the current state and applied torques. The stage cost is captured by $l(\cdot, \cdot, \cdot)$, while $V_f(\cdot, \cdot)$ denotes the terminal cost function. $G(t)$ denotes the teleoperation goal specified by operators. The optimization is performed over a prediction horizon of length $H$. We solve the optimization problem~\eqref{eq:hierarchical_mpc} using kernel interpolation Model Predictive Path Integral (KMPPI) control~\cite{zhong2025rumi}, a sampling-based MPC algorithm that extends Model Predictive Path Integral (MPPI)~\cite{williams2017information} with kernel smoothing for temporally correlated action sequences.

Equations \ref{eq:hmpc_policy} and \ref{eq:hmpc_simulator} together constitute the dynamics model within the MPC framework, enabling the retargeter to regulate the low-level closed-loop dynamics. We adopt MuJoCo as the simulation dynamics $g$ for two reasons. First, prior work has demonstrated that policies evaluated in MuJoCo can achieve zero-shot transfer to real-world hardware \cite{li2025amo, cheng2024expressive}. Second, MuJoCo offers efficient parallel simulation capabilities that are well-suited for integration into MPC-based whole-body control frameworks \cite{alvarez2025real}.

\subsection{State Synchronization}

\begin{table*}[t]
\vspace{2mm}
\centering
\caption{Experimental setups on two platforms. 
In the retargeter commands: 
$h \in \mathbb{R}$: desired waist height; 
$\phi \in \mathbb{R}^3$: desired waist angles; 
$V \in \mathbb{R}^2$: desired base linear velocity (x, y) for the humanoid, 
$V \in \mathbb{R}$: desired base linear velocity for the mobile manipulator; 
$\Omega \in \mathbb{R}$: desired base yaw rate; 
$q_{\text{torso}} \in \mathbb{R}^6$: torso joint positions; 
$q_{\text{leftarm}} \in \mathbb{R}^7$: left-arm joint positions;
$q_{\text{arm}} \in \mathbb{R}^{14}$: joint positions of both arms.}
\label{tab:setup}
\resizebox{\textwidth}{!}{%
\begin{tabular}{lll}
\hline
\textbf{Parameter}                                          & \textbf{Unitree G1 (Humanoid)}                                                             & \textbf{Rainbow RB-Y1 (Mobile Manipulator)}                                                                                                     \\ \hline
\multicolumn{3}{l}{\textit{Robot Configuration}}                                                                                                                                                                                                                                                           \\
DoF; breakdown                                              & 29; 6/leg, 3 waist, 7/arm                                                                  & 22; 2 wheel, 6 torso, 7/arm                                                                                                                     \\
Low-level policy $\pi$ (type); hidden param $h_k$ & FALCON \cite{zhang2025falcon} (RL); state-action history             & Differential drive (proportional control); gain $K_p$ (fixed)                                                                                   \\ \hline
\multicolumn{3}{l}{\textit{High-Level Action $a_k$}}                                                                                                                                                                                                                                                       \\
Retargeter commands (same as inputs to $\pi$)                                         & $h^1, \phi^3, V^2, \Omega^1$ (7-dim, from $T_{\text{head}}$)                               & V1: $V^1, \Omega^1, q_{\text{torso}}^6$ (8-dim, from $T_{\text{head}}$); V2: $q_{\text{leftarm}}^7, q_{\text{torso}}^6$ (13-dim, from $T_{lw}$) \\
IK commands                                                 & $q_{\text{arm}}^{14}$ (from $T_{lw}, T_{rw}$)                                              & V1: $q_{\text{arm}}^{14}$ (from $T_{lw}, T_{rw}$); V2: N/A                                                                                      \\
Mode switch $\sigma$                                        & Standing / Walking (from $\varsigma$)                                                      & Stationary / Mobile (from $\varsigma$)                                                                                                          \\ \hline
\multicolumn{3}{l}{\textit{State Estimation (Real World)}}                                                                                                                                                                                                                                                 \\
$p, \theta, v, \omega$                                      & LiDAR-Inertial SLAM \cite{xu2021fast} ($p,\theta,v$); IMU ($\omega$) & LiDAR-Inertial SLAM \cite{xu2021fast} ($p,\theta,v$); IMU ($\omega$)                                                                                                                              \\
$q, \dot{q}$                                                & Motor encoders                                                                             & Motor encoders                                                                                                                                  \\ \hline
\multicolumn{3}{l}{\textit{MPPI Parameters}}                                                                                                                                                                                                                                                               \\
Horizon $H$ (low-level control step); samples $N$                                    & 3; 30 (Standing), 5; 30 (Walking)                                                          & V1: 3; 30 (Stationary), 10; 30 (Mobile); V2: 3; 50                                                                                              \\
Temperature $\lambda$; noise std $\sigma_{\text{mppi}}$     & 0.001; 0.001 (Standing), 0.001; 0.05 (Walking)                                             & V1: 0.001, 0.01 (Stationary), 0.001, 0.01 (Mobile); V2: 0.001; 0.01                                                                             \\
Control rate (Hz): MPC / low-level policy                       & 25 / 50 (Standing), 10 / 50 (Walking)                                                & 25 / 50 (Stationary), 10 / 50 (Mobile)                                                                                                    \\ \hline
\multicolumn{3}{l}{\textit{Cost Function Weights (Eq. \ref{eq:running_cost})}}                                                                                                                                                                                                      \\
Tracking: $w_p$; $w_\theta$; Energy $w_a$; Contact: $w_s$   & 1.0; 1.0; 0.1; 0.1                                                                         & 1.0; 1.0; 0.1; 0.1                                                                                                                              \\
Collision: $w_c$;\, $w_v$;\, $\alpha$;\, $\beta$            & N/A                                                                                        & 1.0; 1.0; 1.0; 10.0                                                                                                                             \\
Upright constraint (V2 only)                                & N/A                                                                                        & 1.0                                                                                                                                             \\ \hline
Primary safety concern                                      & Balance (measured by ABP)                                                                  & Collision avoidance (measured by CR)                                                                                                            \\ \hline
\end{tabular}
}
\end{table*}

Since our MPC framework treats the closed-loop system comprising the low-level policy and simulation dynamics as its internal model, synchronizing both the hidden parameter $h$ and robot state $x$ between the simulator and real world at each control step is essential. Equations \ref{eq:hmpc_initial_policy} and \ref{eq:hmpc_initial_robot} describe the synchronization of the hidden parameter and robot state, respectively. To enhance generalization, we employ the high-fidelity MuJoCo simulator as the dynamics model rather than deriving explicit analytical equations. Consequently, Equation \ref{eq:hmpc_initial_robot} becomes the problem of synchronizing the MuJoCo robot state with real-world measurements. Given the inherent noise in these measurements, this synchronization becomes a state estimation problem aimed at determining accurate initial conditions for the MuJoCo simulation at each MPC step. This process is particularly critical for legged robots, which are highly sensitive to foot-ground contact conditions.

We formulate the state estimation problem as the following nonlinear least-squares optimization problem:
\begin{align}
\min_{p, \theta, q} \quad & \frac{1}{2} \| p - \tilde{p} \|_{W_p}^2 + \frac{1}{2} \| \theta - \tilde{\theta} \|_{W_\theta}^2 + \frac{1}{2} \| q - \tilde{q} \|_{W_q}^2 \nonumber \\
& + \sum_{i \in \mathcal{C}} \frac{1}{2} \| h_i^{\text{pos}}(p, \theta, q) \|_{W_c^{\text{pos}}}^2 + \sum_{i \in \mathcal{C}} \frac{1}{2} \| h_i^{\text{ori}}(p, \theta, q) \|_{W_c^{\text{ori}}}^2 \nonumber \\
\text{s.t.} \quad & \underline{q} \leq q \leq \overline{q}
\label{eq:state_estimation_qp}
\end{align}

The measurements $\tilde{p}$ and $\tilde{\theta}$ are noisy observations from SLAM algorithm using LiDAR sensors and $\tilde{q}$ is the noisy measurement from motor encoder. $W_p$, $W_\theta$, $W_q$ are positive definite weight matrices reflecting measurement confidence. The set $\mathcal{C}$ contains the indices of active contact points. The functions $h_i^{\text{pos}}(\cdot)$ and $h_i^{\text{ori}}(\cdot)$ represent the contact position and orientation residuals, respectively, with corresponding weight matrices $W_c^{\text{pos}}$ and $W_c^{\text{ori}}$ that penalize constraint violations. $\underline{q}$ and $\overline{q}$ denote the lower and upper joint angle limits, respectively. Notably, the contact constraints are formulated as soft constraints, as jointly satisfying both the measurement alignment and contact constraints is often infeasible.

We denote the position and orientation of the $i$-th contact point in the 
body frame, computed via forward kinematics, as $p_{c,i}(q)$ and $R_{c,i}(q)$, 
respectively. The rotation matrix $R(\theta)$ maps from body to world frame, 
and $e_z = [0,0,1]^\top$ is the vertical unit vector.

\textbf{Humanoid:} The position residual enforces that each foot remains at 
its known stance location $p_{c,i}^{\text{world}}$: 
$h_i^{\text{pos}} = p + R(\theta) \cdot p_{c,i}(q) - p_{c,i}^{\text{world}}$, 
and the orientation residual enforces flat ground contact: 
$h_i^{\text{ori}} = R(\theta) \cdot R_{c,i}(q) \cdot e_z - e_z$. 
We assume rigid foot contact during stance. The active contact set is 
$\mathcal{C} = \{\text{left}, \text{right}\}$ in standing mode, and 
$\mathcal{C} = \{\arg\min_{i} \; e_z^\top (p + R(\theta) \cdot p_{c,i}(q))\}$ 
in walking mode (i.e., only the lower foot).

\textbf{Wheeled robot:} The position residual enforces ground contact at 
wheel radius $r_i$: 
$h_i^{\text{pos}} = e_z^\top (p + R(\theta) \cdot p_{c,i}(q)) - r_i$, 
and the orientation residual enforces that the wheel axle $e_y$ remains 
horizontal: $h_i^{\text{ori}} = e_z^\top (R(\theta) \cdot R_{c,i}(q) \cdot e_y)$. 
The contact set includes all wheels: 
$\mathcal{C} = \{1, \ldots, n_{\text{wheel}}\}$.

We solve the state estimation problem~\eqref{eq:state_estimation_qp} via 
Sequential Quadratic Programming (SQP) using the DAQP solver through 
Mink~\cite{zakka2025mink}.

\subsection{Design of Cost Function}
\label{sec:cost}

The running cost $l(x_k, a_k, G(t))$ is designed to be general across different robotic platforms and consists of four components:
\begin{equation}
\label{eq:running_cost}
\begin{split}
l(x_k, a_k, G(t)) = {} & l_{\text{track}}(x_k, G(t)) + l_{\text{stance}}(x_k) \\
& + l_{\text{col}}(x_k) + l_{\text{ctrl}}(a_k)
\end{split}
\end{equation}

\paragraph{Goal Tracking Cost.} The tracking cost $l_{\text{track}}$ penalizes weighted deviations of the base position $p_k$ and orientation $\theta_k$ from the desired references $p^d(t)$ and $\theta^d(t)$ specified by the teleoperation goal $G(t)$, i.e., $l_{\text{track}} = w_p \|p_k - p^d(t)\|^2 + w_\theta \|\theta_k - \theta^d(t)\|^2$.

\paragraph{Contact Point Cost.} The contact point cost $l_{\text{stance}}$ encourages stability during stance phases by penalizing non-zero velocities $\dot{p}^{c}_{i,k}$ for all active contact points $i \in \mathcal{C}_k$ at each time step, i.e., $l_{\text{stance}} = w_s \sum_{i \in \mathcal{C}_k} \|\dot{p}^{c}_{i,k}\|^2$.

\paragraph{Collision Avoidance Cost.} The collision avoidance cost $l_{\text{col}}$ employs exponential barrier functions over all obstacles, combining a proximity penalty with a velocity-dependent term: $l_{\text{col}} = \sum_{j=1}^{N_{\text{obs}}} \left[ w_c \exp(-\alpha \cdot d_j(x_k)) + w_v \|v_k\|^2 \exp(-\beta \cdot d_j(x_k)) \right]$, where $d_j(x_k)$ is the distance to the $j$-th obstacle, $v_k$ is the robot velocity, and $\alpha, \beta > 0$ control the influence range of each term.

\paragraph{Control Energy Cost.} The control energy cost $l_{\text{ctrl}} = w_a \|a_k\|^2$ penalizes the squared magnitude of high-level actions to encourage energy-efficient motions.

\paragraph{Terminal Cost.} The terminal cost $V_f(x_H, G(t)) = w_f \cdot l_{\text{track}}(x_H, G(t))$ applies a scaled version of the tracking cost at the end of the planning horizon to encourage convergence toward the desired goal state.

\subsection{Input device and input signals}

We employ the Apple Vision Pro (AVP) as the input XR device, leveraging its accurate hand/finger tracking and global headset pose estimation via Visual-Inertial SLAM. The framework uses the headset pose $T_{\text{head}}$, left wrist pose $T_{lw}$, and right wrist pose $T_{rw}$ to control the robot. The flexibility of our proposed retargeter allows users to select any subset of
targets from $G(t)$ to control any subset of body parts based on preference or
task characteristics. For most of robots, the primary embodiment
gap arises from the mobile base. Therefore, in most of the whole-body control tasks, $T_{lw}$ and $T_{rw}$ control
each arm's joint positions $q_{\text{arm}}$ via inverse
kinematics~\cite{zakka2025mink}, while the retargeter focuses on precisely and
safely controlling the base pose ($p$, $\theta$) and non-arm joints
($q_{\text{mobility}}$, $q_{\text{torso}}$) using $T_{\text{head}}$. However,
this is not the only configuration; we also show that for certain tasks, using
only $T_{lw}$ to control all whole-body joints is preferable. A single AVP cannot directly detect foot motions. Prior works infer foot states from fisheye cameras~\cite{wang2024egocentric} or teacher-student frameworks~\cite{li2025clone}, but such indirect measurements introduce substantial uncertainty—e.g., the system may confuse bending with walking intent. We instead use a double-pinch gesture $\varsigma$ to toggle between stationary and mobile modes.

\begin{figure}[t]
    \centering
    \includegraphics[width=\columnwidth]{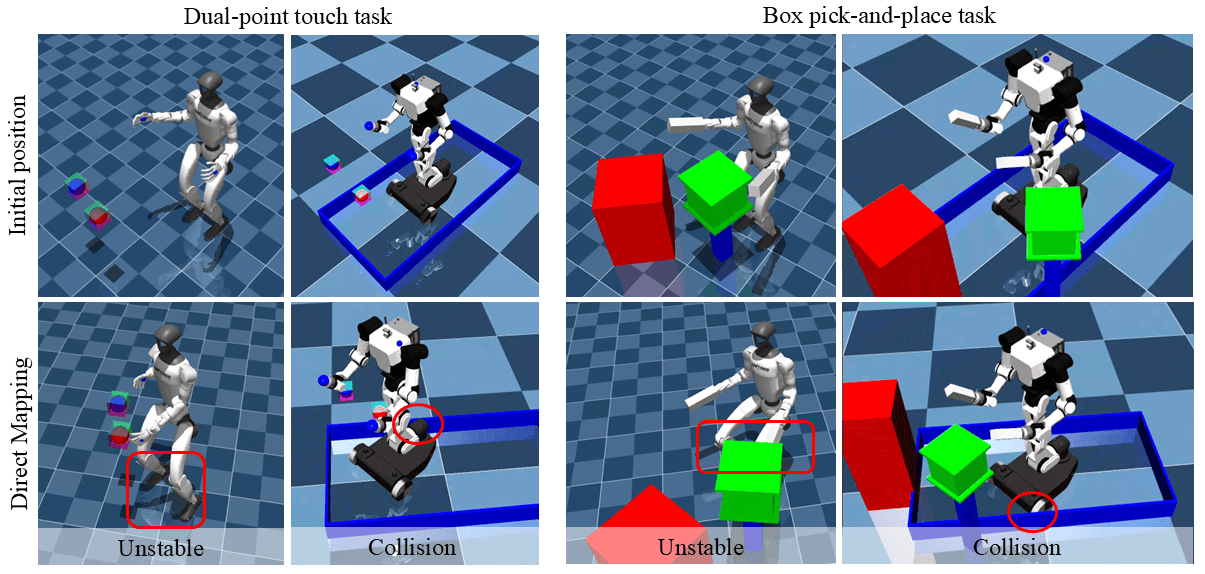}
    \caption{Experimental settings and common failure cases of the direct mapping method. In pick-and-place tasks, the green box is the object and the red box is the table. On the humanoid, unstable upper-body configurations cause loss of balance. On the mobile manipulator, the primary failure mode is collision with surrounding walls (blue boundaries).}
    \label{fig:exp}
\end{figure}

\begin{table*}[t]
\vspace{2mm}
\centering
\caption{Results on \textbf{Humanoid} platform. TSR: Touch Success Rate (\%), PSR: Pickup Success Rate, PPSR: Pick-and-Place Success Rate, CT: Completion Time (s), ABP: Average Balance Power (W).}
\label{tab:humanoid}
\resizebox{\textwidth}{!}{%
\begin{tabular}{llccccccc}
\hline
                 &                            & \multicolumn{3}{c}{\textbf{Dual-point Touch}}                  & \multicolumn{4}{c}{\textbf{Box Pick-and-Place}}                                 \\ \cline{3-9} 
\textbf{Setting} & \textbf{Method}            & TSR $\uparrow$ & CT (s) $\downarrow$    & ABP (W) $\downarrow$ & PSR $\uparrow$ & PPSR $\uparrow$ & CT (s) $\downarrow$   & ABP (W) $\downarrow$ \\
Sim              & Direct mapping with AMO    & 10/10          & 69.62 ± 38.36          & 0.17 ± 0.07          & 10/10          & 1/10            & 49.84 ± 00.00         & 0.28 ± 0.05          \\
                 & Direct mapping with FALCON & 10/10          & 38.30 ± 16.42          & 0.24 ± 0.26          & 8/10           & 1/10            & 78.16 ± 00.00         & 0.57 ± 0.15          \\
                 & Ours w/o state sync        & 10/10          & 52.66 ± 31.44          & 0.43 ± 0.09          & 10/10          & 10/10           & 38.43 ± 8.32          & 1.56 ± 0.04          \\
                 & Ours                       & \textbf{10/10} & \textbf{26.58 ± 6.60}  & \textbf{0.14 ± 0.06} & \textbf{10/10} & \textbf{10/10}  & \textbf{30.75 ± 7.99} & \textbf{0.17 ± 0.05} \\ \hline
Real             & Direct mapping with FALCON & 6/10           & 106.36 ± 35.10         & 0.63 ± 0.06          & 10/10          & 5/10            & 60.34 ± 18.55         & 0.70 ± 0.11          \\
                 & Ours                       & \textbf{10/10} & \textbf{55.83 ± 43.20} & \textbf{0.38 ± 0.08} & \textbf{10/10} & \textbf{9/10}   & \textbf{41.95 ± 3.90} & \textbf{0.39 ± 0.08} \\ \hline
\end{tabular}
}
\end{table*}

\section{Experiments}

This section presents experiments conducted in both simulation and real-world environments on two platforms: the Unitree G1 humanoid and the Rainbow RB-Y1 dual-arm mobile manipulator. The experiments address the following questions:

\begin{itemize} \item How well does the proposed framework perform on various whole-body control (WBC) tasks across the two robotic platforms? \item Compared to other baselines, can the MPC-based retargeter better assist users in teleoperating robots to achieve WBC tasks with greater precision and safety? \item Can the MPC-based retargeter provide users with a flexible and customizable teleoperation experience? \end{itemize}

\subsection{Experimental Tasks}

We select representative WBC tasks from previous research work \cite{zhang2025unleashing} that require coordinated arm movements and base mobility. All experiments presented in this paper were conducted as research experiments in controlled laboratory settings and are not a production deployment. The tasks are defined as follows:

\textbf{Dual-point touch}: Two points (in simulation) or two box holes (in the real world) are sampled in the world space outside the workspace of the robot's initial pose. The operator must reposition the robot such that both points fall within the workspace and then use the arms to pinpoint each target. This task evaluates the precision of the teleoperation system.

\textbf{Box pick-and-place}: A box is initially placed on a support, with a table positioned at a distance of around 1 m. The operator must teleoperate the robot to grasp the box with two arms, transport it to the table by moving the base, and place it on the surface. This task evaluates the overall functionality of the teleoperation system.

\textbf{Non-prehensile transport}: A toy is placed on the open palm without grasping. The user teleoperates the robot to transport the toy to a container with a 10 cm × 5 cm opening, avoiding obstacles along the path while maintaining balance to prevent dropping, and precisely releases it into the hole. This task evaluates the flexibility of the proposed retargeter.

\subsection{Experimental Metrics}

All metrics are reported over 10 trials with methods randomly assigned across trials to mitigate familiarity bias.
\textbf{Touch Success Rate (TSR, \%):} For dual-point touch—percentage of trials where both arms reach within 5\,cm of their targets and hold for 3\,s (simulation), or successfully insert into the holes and lift the box (real world).
\textbf{Pickup / Pick-and-Place Success Rate (PSR / PPSR, \%):} For box pick-and-place—PSR measures successful box lifting (held above support for 3\,s); PPSR measures successful grasping \emph{and} placement on the table without dropping.
\textbf{Delivery Success Rate (DSR, \%):} For non-prehensile transport—percentage of trials where the toy is transported without dropping or hitting obstacles and released into the container.
\textbf{Completion Time (CT, s):} Time to complete each task; for pick-and-place, recorded upon successful table placement.
\textbf{Average Balance Power (ABP, W):} Average motor power of lower-body joints during standing mode, $\text{ABP} = \frac{1}{K}\sum_{k=1}^{K}\frac{1}{m}\sum_{l=1}^{m} P_l(t_k)$, where $K$ is the number of standing-mode time steps, $m$ the number of lower-body joints, and $P_l$ the motor power (torque $\times$ velocity). This quantifies corrective movements and instability.
\textbf{Collision Rate (CR, \%):} Percentage of trials with robot-obstacle collisions.
\textbf{Tilt Angle (TA, rad) and Jerk ($\text{m/s}^3$):} For non-prehensile transport—TA measures average palm deviation from the upright orientation; Jerk measures average palm motion jerk. Both are computed during transport prior to the release motion, reflecting the ability to incorporate additional objectives and tune motion smoothness, respectively.

\subsection{Teleoperation on Humanoid}

\begin{table*}[t]
\vspace{2mm}
\centering
\caption{Results on \textbf{Mobile Manipulator} platform. TSR: Touch Success Rate (\%), PSR: Pickup Success Rate, PPSR: Pick-and-Place Success Rate, CT: Completion Time (s), CR: Collision Rate.}
\label{tab:mobile_manipulator}
\resizebox{\textwidth}{!}{%
\begin{tabular}{llccccccc}
\hline
                 &                               & \multicolumn{3}{c}{\textbf{Dual-point Touch}}                                 & \multicolumn{4}{c}{\textbf{Box Pick-and-Place}}                                                 \\ \cline{3-9} 
\textbf{Setting} & \textbf{Method}               & TSR $\uparrow$ & \multicolumn{1}{l}{CR $\downarrow$} & CT (s) $\downarrow$    & PSR $\uparrow$ & PPSR $\uparrow$ & \multicolumn{1}{l}{CR $\downarrow$} & CT (s) $\downarrow$    \\
Sim              & Direct mapping with diffdrive & 6/10           & 4/10                                & \textbf{21.08 ± 8.89}  & 6/10           & 1/10            & 9/10                                & \textbf{109.28 ± 0.00} \\
                 & Ours                          & \textbf{10/10} & \textbf{0/10}                       & 26.49 ± 13.69          & \textbf{10/10} & \textbf{10/10}  & \textbf{0/10}                       & 139.07 ± 4.32          \\ \hline
Real             & Direct mapping with diffdrive & 9/10           & 1/10                                & \textbf{38.48 ± 10.76} & 10/10          & 1/10            & 5/10                                & \textbf{95.73 ± 0.00}  \\
                 & Ours                          & \textbf{10/10} & \textbf{0/10}                       & 56.75 ± 19.14          & 10/10          & \textbf{7/10}   & \textbf{0/10}                       & 116.5 ± 37.48          \\ \hline
\end{tabular}
}
\end{table*}

\subsubsection{Setup}

We adopt the FALCON RL policy~\cite{zhang2025falcon} 
for force-adaptive loco-manipulation. The retargeter optimizes 7-dim commands 
($h, \phi, V, \Omega$) from $T_{\text{head}}$, while arm joints are set via 
IK from $T_{lw}$ and $T_{rw}$. The detailed setups are listed in Table \ref{tab:setup}.

\subsubsection{Baseline}

To benchmark our MPC-based retargeter, we implement a \emph{direct mapping} baseline following prior work~\cite{li2025amo}, which maps $T_{\text{head}}$ directly to the high-level action space: $h$ is obtained via inverse kinematics using human kinematic models, $\phi$ is derived from the orientation of $T_{head}$, and $V$ and $\Omega$ are computed through finite differencing of $T_{head}$ across consecutive time frames. We test the direct mapping method with both the FALCON~\cite{zhang2025falcon} 
and AMO~\cite{li2025amo} policies. AMO is a whole-body control framework that 
combines RL with trajectory optimization and includes explicit adaptation to 
out-of-distribution (OOD) commands. Both policies share the same input command 
space. Notably, we pair our method only with the FALCON~\cite{zhang2025falcon} policy.

We compare our method against the following baselines:

\textbf{Direct mapping with AMO} \cite{li2025amo}: This baseline demonstrates why the proposed retargeter remains necessary for precise and safe teleoperation, even when the policy itself can handle a certain degree of OOD commands.

\textbf{Direct mapping with FALCON} \cite{zhang2025falcon}: This baseline highlights the capability of our method in handling OOD commands and compensating for imprecise velocity tracking in walking mode. 

\textbf{MPC-based retargeter without state synchronization}: Our MPC-based retargeter using noisy measurements directly from the robot to synchronize the simulated robot at each MPC step.

\subsubsection{Discussion}

Table \ref{tab:humanoid} presents the performance results for two whole-body control tasks—dual-point touch and box pick-and-place—on the humanoid robot in both simulation and real-world settings. The direct mapping methods with AMO and FALCON suffer from several key limitations. First, they exhibit severe drift without a consistent bias, making it difficult for operators to correct; coupled with the inability of human-in-the-loop control to provide the high-frequency adjustments required for precise repositioning during walking, this renders teleoperation intractable and prolongs completion times. Second, because the lower body is controlled by $T_{\text{head}}$ and the operator's head moves continuously, the direct mapping method amplifies unintentional noise, resulting in unstable squatting. Third, even when the robot successfully grasps objects, jerky movements and imprecise velocity tracking cause placement failures—for example, dropping the box or failing to guide the robot near the target table. Fourth, the resulting inefficient and prolonged motions lead to motor overheating, which we observed multiple times in real-world trials. In contrast, our MPC-based retargeter addresses these issues by enabling drift-free repositioning, precisely tracking desired height and waist angles, and regularizing the low-level policy to produce more energy-efficient motions. Although average body power depends on the choice of low-level RL policy—with FALCON producing more aggressive motions than AMO—the retargeter consistently regularizes policy behavior across both, yielding shorter completion times, higher pick-and-place success rates, and reduced energy consumption. Finally, without state synchronization, incorrect foot-ground contact estimation and imprecise future predictions impair the robot's ability to determine optimal actions, substantially degrading performance on the precision-demanding dual-point touch task; the effect is less pronounced for pick-and-place since precise pinpointing is not required, though energy consumption increases due to continuous corrective movements.

\subsection{Teleoperation on Mobile Manipulator}

\subsubsection{Setup}

We adopt a differential drive controller that first converts the desired base velocity $(V, \Omega)$ into desired wheel speeds via kinematic equations, then uses a proportional controller with gain $K_p$ to compute torques that drive the actual wheel speeds toward the desired ones. We define two retargeter variants to demonstrate 
framework flexibility. \textbf{Version~1} mirrors the humanoid setup: 
$T_{\text{head}}$ controls 8-dim actions ($V, \Omega, q_{\text{torso}}$) while 
arms are set via IK. \textbf{Version~2} uses $T_{lw}$ alone to control all 
13 whole-body joints (left arm + torso), with an optional upright constraint 
$h_{\mathrm{ee},L}^{\text{ori}} = R(\theta) \cdot R_{\mathrm{ee},L}(q) \cdot 
e_z - e_z$ added to the cost for non-prehensile transport. This variant 
demonstrates the ease of changing the number of alignment targets, action 
space, and cost function online. In simulation, the robot operates in a walled 
room to test obstacle avoidance in confined environments.
The detailed setups are listed in Table \ref{tab:setup}.

\subsubsection{Baseline}

On this platform, we compare our method against a direct mapping baseline in which $V$ and $\Omega$ are calculated via finite differences across consecutive $T_{head}$ values, and $q_{torso}$ is computed via inverse kinematics from $T_{head}$. Our method can integrate with any low-level policy without retraining for this morphology. Since the AMO and FALCON policies are trained specifically for humanoid robots, combining the direct mapping method with these policies on this platform is non-trivial. Consequently, we compare our method only against the direct mapping method combined with the differential drive controller.

\subsubsection{Discussion}

Table \ref{tab:mobile_manipulator} presents a comparison between our method and the baseline on two whole-body control tasks—dual-point touch and box pick-and-place—for the mobile manipulator. The direct mapping method suffers from two primary limitations. First, the limited field of view from the AVP can occlude obstacles, making it inherently difficult for operators to maneuver safely through the environment; this leads to frequent collisions, particularly in simulation where pinpoint targets are located near walls. Second, these challenges are compounded in the multi-step box pick-and-place task, which requires the robot to navigate toward the box, rotate to avoid collisions, and approach the table without contacting obstacles—resulting in notably low pick and place success rates under direct mapping. While careful human-in-the-loop control can occasionally avoid collisions, the direct mapping results overall reflect the difficulty operators face without active assistance. In contrast, our MPC-based retargeter incorporates a collision avoidance term that continuously repels the robot from obstacles, yielding higher success rates and lower collision rates across both tasks. This collision avoidance behavior does introduce a trade-off: in the simulator, operators require time to adapt to the repulsive forces near walls, resulting in slightly longer completion times for successful trials. In the real world, where maneuvering is considerably easier, both methods achieve high success rates on the dual-point touch task, though our method retains its advantage on the more demanding pick-and-place task. The real-world failure cases for our method resulted from an unstable grasp on the box rather than a navigation or collision issue.

\begin{table}[t]
\vspace{2mm}
\centering
\caption{Non-prehensile transport task on mobile manipulator in real-world. DSR: Delivery Success Rate (\%), CT: Completion Time (s), TA: Tilt Angle (rad), Jerk ($m/s^3$).}
\label{tab:flexibility}
\resizebox{\columnwidth}{!}{%
\begin{tabular}{lcccc}
\hline
\textbf{Setting}  & DSR $\uparrow$ & CT (s) $\downarrow$ & TA (rad) $\downarrow$ & Jerk ($m/s^3$) $\downarrow$ \\ \hline
\makecell[l]{3-Point Alignment\\w/o Upright Constraint} & 1/10          & 30.80 ± 0.00    & 0.41 ± 0.12     & 1259.51 ± 256.05  \\ \hline
\makecell[l]{1-Point Alignment\\w/ Upright Constraint} & \textbf{10/10}         & 34.50 ± 6.37    & 0.03 ± 0.03     & 1059.44 ± 191.45  \\ \hline
Conservative      & 7/10          & 40.75 ± 1.06    & 0.05 ± 0.03     & \textbf{769.99 ± 11.02}    \\ \hline
Aggressive        & 6/10          & \textbf{24.50 ± 3.50}    & \textbf{0.01 ± 0.01}     & 2144.55 ± 87.80   \\ \hline
\end{tabular}
}
\end{table}

Table~\ref{tab:flexibility} demonstrates the flexibility of our framework for 
a real-world non-prehensile transport task on the mobile manipulator. We first 
vary the weight of the control energy cost (Section~\ref{sec:cost}) to adjust 
the conservativeness of the motion. A larger weight ($w_a = 0.5$) yields 
smoother, more conservative motion (lower jerk) at the expense of longer 
completion time; however, the slow response introduces delay, leading to 
collisions with the surrounding environment. Conversely, a smaller weight 
($w_a = 0.02$) produces more aggressive and responsive motion with shorter 
completion time but higher jerk. Failures in this setting correspond to the 
toy being dropped during transport due to the jerky motions. We further 
demonstrate the ease of changing the number of alignment targets and the cost 
function online. In the 3-point alignment setting, no upright constraint is 
applied, requiring the user to manually keep the end-effector level based on 
experience. In the 1-point alignment setting, an upright constraint is imposed 
on the end-effector during transport; once the robot reaches the container 
opening, the user removes this constraint online via a gesture to perform the 
releasing motion. The results show that incorporating the upright constraint 
significantly improves the success rate, as evidenced by smaller tilt angle 
(TA) and lower jerk values during transport.

\section{Conclusion}

We presented a hierarchical whole-body teleoperation framework that consists of 
an MPC-based motion retargeter to control diverse robot morphologies from a 
single XR device without retraining low-level policies. A state synchronization 
method ensures robust online execution under noisy measurements and contact 
sensitivity. Experiments on a Unitree G1 humanoid and a Rainbow RB-Y1 mobile 
manipulator demonstrated higher success rates with improved precision and safety 
over direct mapping baselines. As a plug-and-play module compatible with 
arbitrary controllers and platforms, our framework offers a scalable, 
morphology-agnostic solution for whole-body teleoperation. Looking ahead, we 
plan to improve control rate and responsiveness by distilling the MPC policy 
into a neural network for real-time execution, and to enable compliant behavior 
during contact-rich tasks by integrating force sensing into the framework.

\section*{ACKNOWLEDGMENT}

The authors would like to thank Haowen Liu, Bharath Masetty, and Yinai Fan for their assistance with the setup of experiments, Bill Smart, Roberto Martin-Martin, Taskin Padir, and Shang Meng for their review, and our colleagues at Amazon for their support. Jen-Wei Wang thanks Shao-Yi Yu for her constant encouragement during this project.

\bibliographystyle{IEEEtran}
\bibliography{IEEEabrv, references}

\begin{thebibliography}{10}
\providecommand{\url}[1]{#1}
\csname url@samestyle\endcsname
\providecommand{\newblock}{\relax}
\providecommand{\bibinfo}[2]{#2}
\providecommand{\BIBentrySTDinterwordspacing}{\spaceskip=0pt\relax}
\providecommand{\BIBentryALTinterwordstretchfactor}{4}
\providecommand{\BIBentryALTinterwordspacing}{\spaceskip=\fontdimen2\font plus
\BIBentryALTinterwordstretchfactor\fontdimen3\font minus
  \fontdimen4\font\relax}
\providecommand{\BIBforeignlanguage}[2]{{%
\expandafter\ifx\csname l@#1\endcsname\relax
\typeout{** WARNING: IEEEtran.bst: No hyphenation pattern has been}%
\typeout{** loaded for the language `#1'. Using the pattern for}%
\typeout{** the default language instead.}%
\else
\language=\csname l@#1\endcsname
\fi
#2}}
\providecommand{\BIBdecl}{\relax}
\BIBdecl

\bibitem{ben2025homie}
Q.~Ben, F.~Jia, J.~Zeng, J.~Dong, D.~Lin, and J.~Pang, ``Homie: Humanoid
  loco-manipulation with isomorphic exoskeleton cockpit,'' \emph{arXiv preprint
  arXiv:2502.13013}, 2025.

\bibitem{ze2025twist}
Y.~Ze, Z.~Chen, J.~P. Ara{\'u}jo, Z.-a. Cao, X.~B. Peng, J.~Wu, and C.~K. Liu,
  ``Twist: Teleoperated whole-body imitation system,'' \emph{arXiv preprint
  arXiv:2505.02833}, 2025.

\bibitem{li2025clone}
Y.~Li, Y.~Lin, J.~Cui, T.~Liu, W.~Liang, Y.~Zhu, and S.~Huang, ``Clone:
  Closed-loop whole-body humanoid teleoperation for long-horizon tasks,''
  \emph{arXiv preprint arXiv:2506.08931}, 2025.

\bibitem{li2025amo}
J.~Li, X.~Cheng, T.~Huang, S.~Yang, R.-Z. Qiu, and X.~Wang, ``Amo: Adaptive
  motion optimization for hyper-dexterous humanoid whole-body control,''
  \emph{arXiv preprint arXiv:2505.03738}, 2025.

\bibitem{seo2025legato}
M.~Seo, H.~A. Park, S.~Yuan, Y.~Zhu, and L.~Sentis, ``Legato: Cross-embodiment
  imitation using a grasping tool,'' \emph{IEEE Robotics and Automation
  Letters}, 2025.

\bibitem{mahmood2019amass}
N.~Mahmood, N.~Ghorbani, N.~F. Troje, G.~Pons-Moll, and M.~J. Black, ``Amass:
  Archive of motion capture as surface shapes,'' in \emph{Proceedings of the
  IEEE/CVF international conference on computer vision}, 2019, pp. 5442--5451.

\bibitem{darvish2019whole}
K.~Darvish, Y.~Tirupachuri, G.~Romualdi, L.~Rapetti, D.~Ferigo, F.~J.~A.
  Chavez, and D.~Pucci, ``Whole-body geometric retargeting for humanoid
  robots,'' in \emph{2019 IEEE-RAS 19th International Conference on Humanoid
  Robots (Humanoids)}.\hskip 1em plus 0.5em minus 0.4em\relax IEEE, 2019, pp.
  679--686.

\bibitem{araujo2025retargeting}
J.~P. Araujo, Y.~Ze, P.~Xu, J.~Wu, and C.~K. Liu, ``Retargeting matters:
  General motion retargeting for humanoid motion tracking,'' \emph{arXiv
  preprint arXiv:2510.02252}, 2025.

\bibitem{zhang2025falcon}
Y.~Zhang, Y.~Yuan, P.~Gurunath, I.~Gupta, S.~Omidshafiei, A.-a. Agha-mohammadi,
  M.~Vazquez-Chanlatte, L.~Pedersen, T.~He, and G.~Shi, ``Falcon: Learning
  force-adaptive humanoid loco-manipulation,'' \emph{arXiv preprint
  arXiv:2505.06776}, 2025.

\bibitem{williams2017information}
G.~Williams, N.~Wagener, B.~Goldfain, P.~Drews, J.~M. Rehg, B.~Boots, and E.~A.
  Theodorou, ``Information theoretic mpc for model-based reinforcement
  learning,'' in \emph{2017 IEEE international conference on robotics and
  automation (ICRA)}.\hskip 1em plus 0.5em minus 0.4em\relax IEEE, 2017, pp.
  1714--1721.

\bibitem{lu2025mobile}
C.~Lu, X.~Cheng, J.~Li, S.~Yang, M.~Ji, C.~Yuan, G.~Yang, S.~Yi, and X.~Wang,
  ``Mobile-television: Predictive motion priors for humanoid whole-body
  control,'' in \emph{2025 IEEE International Conference on Robotics and
  Automation (ICRA)}.\hskip 1em plus 0.5em minus 0.4em\relax IEEE, 2025, pp.
  5364--5371.

\bibitem{he2025asap}
T.~He, J.~Gao, W.~Xiao, Y.~Zhang, Z.~Wang, J.~Wang, Z.~Luo, G.~He, N.~Sobanbab,
  C.~Pan \emph{et~al.}, ``Asap: Aligning simulation and real-world physics for
  learning agile humanoid whole-body skills,'' \emph{arXiv preprint
  arXiv:2502.01143}, 2025.

\bibitem{he2024learning}
T.~He, Z.~Luo, W.~Xiao, C.~Zhang, K.~Kitani, C.~Liu, and G.~Shi, ``Learning
  human-to-humanoid real-time whole-body teleoperation,'' in \emph{2024
  IEEE/RSJ International Conference on Intelligent Robots and Systems
  (IROS)}.\hskip 1em plus 0.5em minus 0.4em\relax IEEE, 2024, pp. 8944--8951.

\bibitem{he2024omnih2o}
T.~He, Z.~Luo, X.~He, W.~Xiao, C.~Zhang, W.~Zhang, K.~Kitani, C.~Liu, and
  G.~Shi, ``Omnih2o: Universal and dexterous human-to-humanoid whole-body
  teleoperation and learning,'' \emph{arXiv preprint arXiv:2406.08858}, 2024.

\bibitem{cheng2024expressive}
X.~Cheng, Y.~Ji, J.~Chen, R.~Yang, G.~Yang, and X.~Wang, ``Expressive
  whole-body control for humanoid robots,'' \emph{arXiv preprint
  arXiv:2402.16796}, 2024.

\bibitem{zhong2025rumi}
S.~Zhong, N.~Fazeli, and D.~Berenson, ``Rumi: Rummaging using mutual
  information,'' \emph{IEEE Transactions on Robotics}, 2025.

\bibitem{alvarez2025real}
J.~Alvarez-Padilla, J.~Z. Zhang, S.~Kwok, J.~M. Dolan, and Z.~Manchester,
  ``Real-time whole-body control of legged robots with model-predictive path
  integral control,'' in \emph{2025 IEEE International Conference on Robotics
  and Automation (ICRA)}.\hskip 1em plus 0.5em minus 0.4em\relax IEEE, 2025,
  pp. 14\,721--14\,727.

\bibitem{xu2021fast}
W.~Xu and F.~Zhang, ``Fast-lio: A fast, robust lidar-inertial odometry package
  by tightly-coupled iterated kalman filter,'' \emph{IEEE Robotics and
  Automation Letters}, vol.~6, no.~2, pp. 3317--3324, 2021.

\bibitem{zakka2025mink}
K.~Zakka, ``Mink: Python inverse kinematics based on mujoco,'' 2025.

\bibitem{wang2024egocentric}
J.~Wang, Z.~Cao, D.~Luvizon, L.~Liu, K.~Sarkar, D.~Tang, T.~Beeler, and
  C.~Theobalt, ``Egocentric whole-body motion capture with fisheyevit and
  diffusion-based motion refinement,'' in \emph{Proceedings of the IEEE/CVF
  Conference on Computer Vision and Pattern Recognition}, 2024, pp. 777--787.

\bibitem{zhang2025unleashing}
Z.~Zhang, C.~Chen, H.~Xue, J.~Wang, S.~Liang, Y.~Liu, Z.~Zhang, H.~Wang, and
  L.~Yi, ``Unleashing humanoid reaching potential via real-world-ready skill
  space,'' \emph{IEEE Robotics and Automation Letters}, vol.~11, no.~2, pp.
  2082--2089, 2025.

\end{thebibliography}

\end{document}